\documentclass[letterpaper, 10 pt, conference]{ieeeconf}  

\IEEEoverridecommandlockouts                              

\usepackage{amsmath} 
\usepackage{amssymb}  

\usepackage{graphicx}
\usepackage{tabu}
\usepackage{diagbox}
\usepackage{xcolor}
\usepackage{soul}
\usepackage{caption}
\title{\LARGE \bf
Robustness of Utilizing Feedback in Embodied Visual Navigation
}

\author{Jenny Zhang$^{1}$, Samson Yu$^{2}$, Jiafei Duan$^{3}$ and Cheston Tan$^{4}$%
\thanks{$^{1}$Author is with University of British Columbia, Canada, {\tt\small jennyzzt@cs.ubc.ca}}%
\thanks{$^{2}$Author is with National University of Singapore, Singapore}%
\thanks{$^{3}$Author is with University of Washington, USA}%
\thanks{$^{4}$Author is with Centre for Frontier AI Research, A*STAR, Singapore}%
}

\begin{document}

\maketitle
\thispagestyle{empty}
\pagestyle{empty}

\begin{abstract}
This paper presents a framework for training an agent to actively request help in object-goal navigation tasks, with feedback indicating the location of the target object in its field of view. To make the agent more robust in scenarios where a teacher may not always be available, the proposed training curriculum includes a mix of episodes with and without feedback. The results show that this approach improves the agent's performance, even in the absence of feedback.
\end{abstract}

\vspace{-1mm}
\section{Introduction}
\begin{figure}[ht]
    \centering
    \includegraphics[width=1\linewidth]{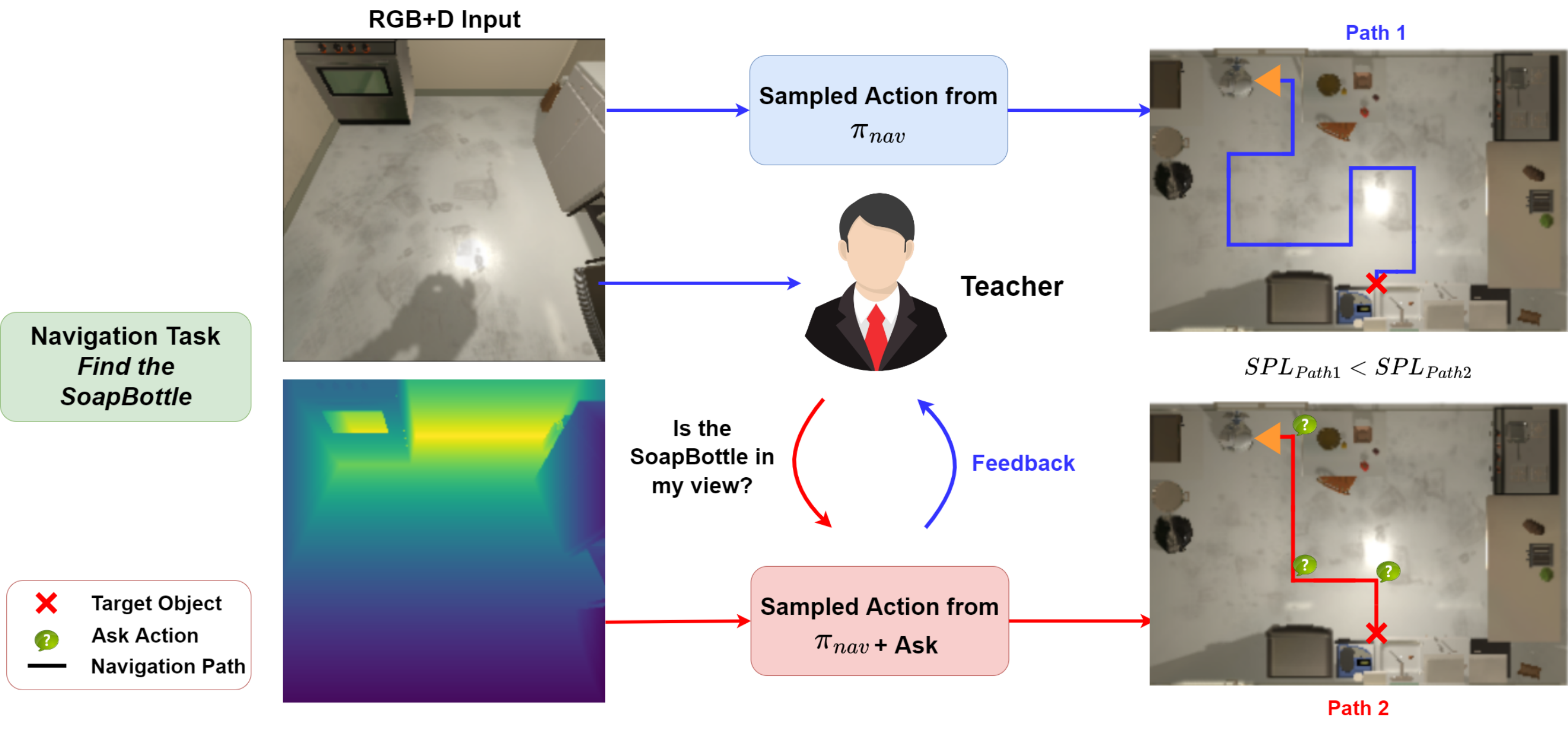}
    \caption{Our learning framework with \emph{ask}-for-feedback capability in ObjectNav. A comparison between agents' performance with and without the \emph{ask} capability.}
\label{fig:0}
\end{figure}

A newly deployed robot assistant struggles to retrieve a tool, wasting time searching the entire house (top-right of Figure \ref{fig:0}). It could have asked for help from someone familiar with the environment. To make robots more effective, they need the ability to actively acquire relevant information \cite{chi2020just,chen2011learning,nguyen2021learning}. This paper proposes a learning framework to address this challenge.

\vspace{-2mm}

\vspace{-5mm}
\section{Methods}
We tackle the ObjectNav task \cite{batra2020objectnav,eaisurvey,anderson2018vision} in AI2-THOR \cite{kolve2017ai2} environment. The goal is to navigate to a target object and issue a termination action while maintaining a stopping distance of 1.0m or less. Two key differences with \cite{batra2020objectnav} is that the target object is randomly chosen, and the placement of objects is randomized every episode.

The agent has six basic navigation actions \cite{batra2020objectnav} and, if feedback is enabled, an additional action of asking for help. The agent is equipped with a RGB-D sensor. Upon each \textit{ask} action, an \textbf{object-in-view} observation is provided, which includes the ground truth semantic segmentation observation of the target object's location in the agent's view. Pixels corresponding to the target object's location have a value of 1, and 0 for the rest.

In real-world settings, the teacher may not always be present to provide assistance. We use a semi-present teacher training curriculum to improve the agent's robustness in such settings. Two curricula are compared: 25\% and 75\% semi-present teacher. In a $\eta$\% semi-present teacher curriculum, the teacher is present in $\eta$\% of training episodes. When the teacher present, the agent receives an additional observation, and feedback is only available when there is a teacher.

\section{Experimental Setup \& Results}
The agent is trained using the PPO \cite{ppo} reinforcement learning algorithm and the Adam \cite{adamoptimizer} optimization algorithm. The performance of the agent is evaluated using two standard navigation metrics \cite{batra2020objectnav}: \textbf{success rate (SR)} and \textbf{success weighted by path length (SPL)}.

\vspace{-2mm}
\begin{table}[ht]
\centering
\tabulinesep=0mm
\tabcolsep=0.5mm
\begin{tabu}{clccccc}
\hline
 & \multicolumn{1}{c}{} & \multicolumn{2}{c}{SR (\%)} & \multicolumn{1}{l}{} & \multicolumn{2}{c}{SPL (\%)} \\ \cline{3-4} \cline{6-7} 
\begin{tabular}[c]{@{}c@{}}Teacher Presence\\during testing\end{tabular} & \begin{tabular}[c]{@{}c@{}}Training\\Methods\end{tabular} & \begin{tabular}[c]{@{}c@{}}Seen\\Objects\end{tabular}& \begin{tabular}[c]{@{}c@{}}Unseen\\Objects\end{tabular} & \multicolumn{1}{l}{} & \begin{tabular}[c]{@{}c@{}}Seen\\Objects\end{tabular}& \begin{tabular}[c]{@{}c@{}}Unseen\\Objects\end{tabular} \\ \hline
False & Baseline & 35.2 & 13.0 &  & 24.1 & 6.9 \\
False & Semi-25 & 40.6 & 13.3 &  & 26.8 & 18.5 \\
False & Semi-75 & 33.9 & 18.5 &  & 27.4 & 16.3 \\ \hline
True & Feedback & 70.9 & \textbf{37.0} &  & 46.4 & \textbf{24.2} \\
True & Semi-25 & 51.6 & 15.1 &  & 34.1 & 8.3 \\
True & Semi-75 & \textbf{72.1} & 24.8 &  & \textbf{50.9} & 18.9 \\ \hline
\end{tabu}%
\caption{Success Rate (SR) and Success weighted by Path Length (SPL) evaluation results for seen and unseen objects.
}
\label{tab:results-basic}
\vspace{-1mm}
\end{table}

Table \ref{tab:results-basic} shows that the object-in-view feedback method outperforms the baseline, resulting in an average improvement of 29.9\% for SR and 19.8\% for SPL. As expected, the agent's performance decreases when objects are unseen during training.
The agent trained with object-in-view feedback and a 100\% present teacher performed poorly when the teacher was absent (95\% failure rate), compared to the baseline which had a success rate of at least 13\%. However, the agents trained with curricula that included 25\% or 75\% teacher presence performed similarly to the baseline when the teacher was absent, and better than the baseline when feedback was available. This indicates that a mix of episodes with and without teacher feedback enables the agent become more robust in the absence of the teacher.

\section{Conclusion}
This work highlights the benefits of incorporating feedback into an agent's navigation strategy. Our results show the effectiveness of a training curriculum that includes instances where feedback is not always available. Interesting future works include exploring alternative feedback modalities and developing methods to quantify the agent's uncertainty about the location of the goal.

\bibliography{references}

\end{document}